%% file: iclr2024_conference.tex
\documentclass{article} 
\usepackage{iclr2024_conference,times}

\input{math_commands.tex}

\usepackage{hyperref}
\hypersetup{
	colorlinks=true,
	linkcolor=red,
	filecolor=blue,      
	urlcolor=magenta,
	citecolor=black,
}
\usepackage{graphicx}
\usepackage{multirow}  
\usepackage{booktabs}  
\usepackage{wrapfig}
\usepackage{xspace}
\usepackage{url}
\newcommand{\method}{\textsc{SemanticSDS}\xspace}

\title{Semantic Score Distillation Sampling for Compositional Text-to-3D Generation}

\author{\textbf{Ling Yang{$^{1\dag}$}\thanks{Contributed equally.},\ Zixiang Zhang{$^{1}$}\footnotemark[1], \ Junlin Han {$^{2}$},\ Bohan Zeng{$^{1}$},\ Runjia Li{$^{2}$}}\\ \textbf{Philip Torr{$^{2}$}, \ \textbf{Wentao Zhang}{$^{1}$}\thanks{Corresponding authors: yangling0818@163.com, wentao.zhang@pku.edu.cn}} \\ 
    {$^1$}Peking University \quad {$^2$}University of Oxford\\
Project: \href{https://github.com/YangLing0818/SemanticSDS-3D}{https://github.com/YangLing0818/SemanticSDS-3D}
}

\iclrfinalcopy 
\begin{document}

\maketitle

\begin{abstract}
Generating high-quality 3D assets from textual descriptions remains a pivotal challenge in computer graphics and vision research. Due to the scarcity of 3D data, state-of-the-art approaches utilize pre-trained 2D diffusion priors, optimized through Score Distillation Sampling (SDS).
Despite progress, crafting complex 3D scenes featuring multiple objects or intricate interactions is still difficult. To tackle this, recent methods have incorporated box or layout guidance.
However, these layout-guided compositional methods often struggle to provide fine-grained control, as they are generally coarse and lack expressiveness.
To overcome these challenges, we introduce a novel SDS approach, Semantic Score Distillation Sampling (\method), designed to effectively improve the expressiveness and accuracy of compositional text-to-3D generation.
Our approach integrates new semantic embeddings that maintain consistency across different rendering views and clearly differentiate between various objects and parts.
These embeddings are transformed into a semantic map, which directs a region-specific SDS process, enabling precise optimization and compositional generation. By leveraging explicit semantic guidance, our method unlocks the compositional capabilities of existing pre-trained diffusion models, thereby achieving superior quality in 3D content generation, particularly for complex objects and scenes.
Experimental results demonstrate that our \method framework is highly effective for generating state-of-the-art complex 3D content.
\end{abstract}

\section{Introduction}
Generating high-quality 3D assets from textual descriptions is a long-standing
goal in computer graphics and vision research.
However, due to the scarcity of 3D data, existing text-to-3D generation models have primarily relied on leveraging powerful pre-trained 2D diffusion priors to optimize 3D representations, typically based on a score distillation sampling (SDS) loss~\citep{poole2023dreamfusion}.
Notable examples include DreamFusion, which pioneered the use of SDS to optimize Neural Radiance Field (NeRF) representations~\citep{mildenhall2021nerf}, and Magic3D~\citep{lin2023magic3d}, which further advanced this approach by proposing a coarse-to-fine framework to enhance its performance.

\begin{figure}[t]
  \centering
  \vspace{-0.4in}
  \includegraphics[width=0.99\linewidth]{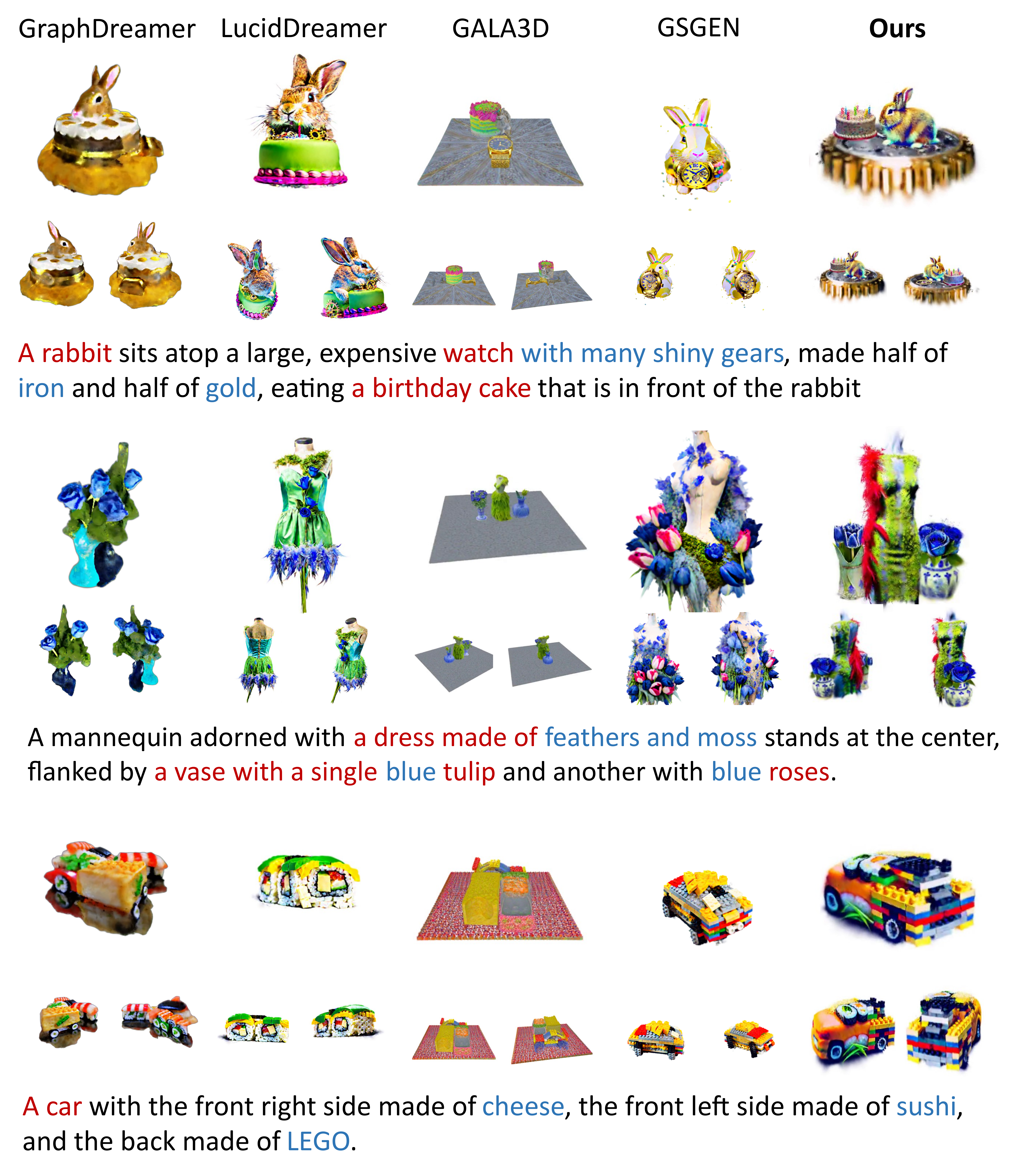}
  \caption{\method achieves superior compositional text-to-3d generation results over state-of-the-art baselines, particularly in generating multiple objects with diverse attibutes.}
  \label{fig:teaser}
  \vspace{-4mm}
\end{figure}

Despite the advancements in lifting and SDS-based methods, generating complex 3D scenes with multiple objects or intricate interactions remains a significant challenge. Recent efforts have focused on incorporating additional guidance, such as box or layout information\citep{po2024compositional,epstein2024disentangled,zhou2024gala3d}.
Among them, \citet{po2024compositional} introduce locally conditioned
diffusion for compositional scene diffusion based on input bounding boxes with one shared NeRF representation while \citet{epstein2024disentangled} instantiate and render multiple NeRFs
for a given scene using each NeRF to represent a separate 3D entity with a set of layouts. 
Further advancing this field, GALA3D~\citep{zhou2024gala3d} utilizes large language models (LLMs) to generate coarse layouts to guide 3D generation for compositional scenes.

However, existing layout-guided compositional methods often fall short in achieving fine-grained control over the generated 3D scenes. The current form of box or layout guidance is relatively coarse and lacks the expressiveness required to effectively guide the SDS process in optimizing the intricate interactions or intersecting parts between multiple objects, particularly when generating objects with multiple attributes.
This limitation stems from the fact that pre-trained 2D diffusion models, which are used in SDS, struggle to estimate accurate scores for complex scenarios with consistent views when explicit spatial guidance is absent~\citep{li2023gligen, shi2024mvdream}. As a result, the generated 3D scenes may lack the level of detail and realism desired, highlighting the need for more precise guidance mechanisms that can provide finer-grained control over the generation process.

To address these limitations, we propose Semantic Score Distillation Sampling (\method), which boosts the expressiveness and precision of compositional text-to-3D generation. For more explicit 3D expression, we equip \method with 3D Gaussian Splatting (3DGS)~\citep{kerbl20233d} as the 3D representation.
Our approach consists of three key steps:
(1) Given a text prompt, we propose a program-aided approach to improve the accuracy of LLM-based layout planning for 3D scenes.
(2) We introduce novel semantic embeddings that remain consistent across various rendering views and explicitly distinguish different objects and parts.
(3) We then render these semantic embeddings into a semantic map, which serves as guidance for a region-wise SDS process, facilitating fine-grained optimization and compositional generation.
Our approach addresses the challenge of leveraging pre-trained diffusion models, which possess powerful compositional diffusion priors but are difficult to utilize \citep{wang2024compositional,yang2024mastering}. By using explicit semantic map guidance, we innovatively unlock these compositional 2D diffusion priors for high-quality 3D content generation.

Our main contributions are summarized as follows:
\begin{itemize}

\item We propose \method, a novel semantic-guided score distillation sampling approach that effectively enhances the expressiveness and precision of compositional text-to-3D generation,  as shown in Figure~\ref{fig:teaser}.
\item We introduce program-aided layout planning to improve positional and relational accuracy in generated 3D scenes, deriving precise 3D coordinates from ambiguous descriptions.
\item We develop expressive semantic embeddings to augment 3D Gaussian representations, and propose a region-wise SDS process with the rendered semantic map,
distinguishing different objects and parts in the compositional generation process.
\end{itemize}

\section{Related Work}
\paragraph{Text-to-3D Generation}
Different approaches have been developed to achieve text-to-3D content generation \citep{deitke2024objaverse,zeng2023ipdreamer}, such as employing multi-view diffusion models \citep{shi2024mvdream,wu2024unique3d,kong2024eschernet,blattmann2023stable}, direct 3D diffusion models \citep{gupta20233dgen,shue20233d,wu2024direct3d} and large reconstruction models \citep{hong2023lrm}. 
For instance, multi-view diffusion models are trained and optimized by fine-tuning video diffusion on 3D datasets, aiding in 3D reconstruction \citep{voleti2024sv3d,chen2024v3d,han2024vfusion3d}. \citet{you2024nvs} propose a training-free method that employs video diffusion as a zero-shot novel view synthesizer. 
However, these methods require numerous 3D data for training. In contrast, Score Distillation Sampling (SDS) ~\citep{poole2023dreamfusion, wang2023score} is 3D data-free and generally produces higher quality assets. 
SDS approaches harness the creative potential of 2D diffusion and have achieved significant advancements ~\citep{wang2024prolificdreamer, yang2023learn, hertz2023delta}, resulting in realistic 3D content generation and enhanced resolution of generative models \citep{zhu2024hifa}.
In this paper, we propose a new SDS paradigm, namely \method, for text-to-3D generation in complex scenarios, which first incorporates explicit semantic guidance into the SDS process.

\paragraph{Compositional 3D Generation}
Modeling compositional 3D data distribution is a fundamental and critical task for generative models. 
Current feed-forward methods \citep{shue20233d,shi2024mvdream}  are primarily capable of generating single objects and face challenges when creating more complex scenes containing multiple objects due to limited training data. \citet{po2024compositional} fix the layout in multiple 3D bounding boxes and generate compositional assets with bounding-box-specific SDS.
Recently, a series of learnable-layout compositional methods have been proposed \citep{epstein2024disentangled,vilesov2023cg3d,han2024reparo,chen2024comboverse, li2024dreambeastdistilling3dfantastical,yan2024frankenstein,gao2024graphdreamer} . These methods combine multiple object-ad-hoc radiance fields and then optimize the positions of the radiance fields from external feedback. For example,
\citet{epstein2024disentangled} propose learning a distribution of reasonable layouts based solely on the knowledge from a large pre-trained text-to-image model.   
\citet{vilesov2023cg3d} introduce an optimization method based on Monte-Carlo sampling and physical constraints. 
Non-learnable layout methods like \citep{zhou2024gala3d} and \citet{lin2023towards} further utilize LLMs or MLLMs to convert text into reasonable layouts. 
However, the current form of layout guidance is relatively coarse and not expressive enough for fine-grained control. We address this problem by incorporating semantic embeddings that ensure view consistency and distinctly differentiate objects into SDS processes, which are flexible and expressive for optimizing 3D scenes.

\section{Preliminaries}

\paragraph{Compositional 3D Gaussian Splatting} 
3D Gaussian Splatting explicitly represents a 3D scene as a collection of anisotropic 3D Gaussians, each characterized by a mean $\mu \in \mathbb{R}^3$ and a covariance matrix $\Sigma$~\citep{kerbl20233d}. The Gaussian function $G(x)$ is defined as:
\begin{equation}
    G(x) = \exp \left(- \frac{1}{2} (x - \mu)^{\top} \Sigma^{-1} (x - \mu)\right)
    \label{eq:3dGaussian}
\end{equation}

Rendering a compositional scene necessitates a transformation from object to composition coordinates, involving a rotation $\mathbf{R} \in \mathbb{R}^{3 \times 3}$, translation $\mathbf{t} \in \mathbb{R}^3$, and scale $s \in \mathbb{R}$~\citep{zhou2024gala3d,vilesov2023cg3d}. This transformation is applied to the mean and variance of individual Gaussians, transitioning from the object's local coordinates to global coordinates:
$
\mu^{\mathrm{global}} = s \mathbf{R} \mu^{\mathrm{local}} + \mathbf{t}$,
$
\mathbf{\Sigma^{\mathrm{global}}} = s^{2} \mathbf{R} \mathbf{\Sigma}^{\mathrm{local}} \mathbf{R}^{\top}
$.

For optimized rendering of compositional 3D Gaussians into 2D image planes, a tile-based rasterizer enhances rendering efficiency. The rendered color at pixel $v$ is computed as follows:
\begin{equation}
    \mathbf{I}(v) = \sum_{i \in \mathcal{N}} c_i \alpha_i \prod_{j=1}^{i-1} (1 - \alpha_j),
    \label{eq:rendering_3dgs}
\end{equation}
where $c_i$ represents the color of the $i$-th Gaussian, $\mathcal{N}$ denotes the set of Gaussians within the tile, and $\alpha_i$ is the opacity.

\paragraph{Score Distillation Sampling}
\citet{yang2023diffusion, wang2023score} have introduced a method to leverage a pretrained diffusion model, $\epsilon_{\phi}(x_t; y, t)$, to optimize the 3D representation, where $x_t$, $y$, and $t$ signify the noisy image, text embedding, and timestep, respectively.

Let $g$ represent the differentiable rendering fcuntion, $\theta$ denote the parameters of the optimizable 3D representation and $\mathbf{I} = g(\theta)$ be the resulting rendered image. The gradient for optimization is performed via Score Distillation Sampling:
\begin{equation}
\nabla_\theta \mathcal{L}_{\mathrm{SDS}}=\mathbb{E}_{\epsilon, t}\left[w(t)\left(\epsilon_\phi\left(x_t ; y, t\right)-\epsilon\right) \frac{\partial \mathbf{I}}{\partial \theta}\right]
\end{equation}
where $\epsilon$ is Gaussian noise and $w(t)$ is a weighting function. In compositional 3D generation, local object optimizations and global scene optimizations alternate in a compositional optimization scheme~\citep{zhou2024gala3d}. During local optimization, the parameters \(\theta\) include the mean, covariance, and color of individual Gaussians. In global scene optimization, the parameters \(\theta\) additionally include transformations—translation, scale, and rotation—that convert local to global coordinates.

\section{Method}

\begin{figure}[t]
  \vspace{-0.3in}
  \centering
  \includegraphics[width=0.98\linewidth]{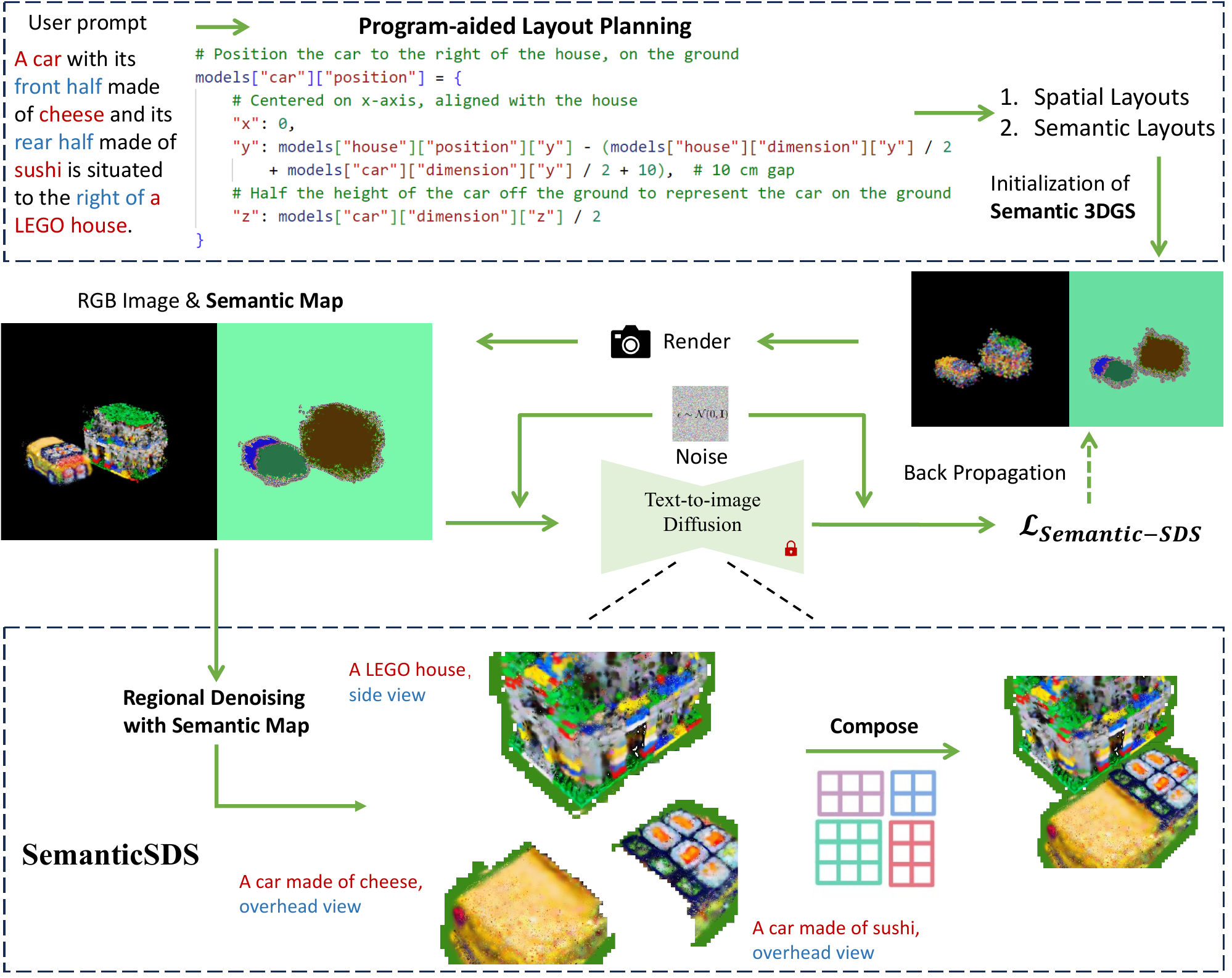}
  \caption{Overview of \method, comprising of program-aided layout planning (top) and regional denoising with semantic map (bottom).}
  \label{fig:framework}
\end{figure}

\subsection{Program-aided Layout Planning}
\label{subsection:program_aided}

A detailed characterization of multiple objects' positions, dimensions, and orientations requires numerous parameters, especially when additionally describing distinct attributes of various object components. In scenarios involving multiple objects, utilizing Large Language Models (LLMs) to derive precise 3D coordinates from ambiguous descriptions within a scene is often challenging. This difficulty arises because purely 3D numerical data and corresponding natural language descriptions do not frequently co-occur in the training data of LLMs~\citep{hong20233d, xu2023pointllm}. Consequently, issues such as overlapping objects or excessive distances between them may occur, particularly during interactions among objects. Therefore, we propose to leverage programs as the intermediate reasoning and planning steps ~\citep{gao2023pal} to effectively mitigate these challenges.

Let $y_c$ represent the complex user input, which includes multiple objects with various attributes. First, We utilize Large Language Models to identify all objects $\left\{O_k\right\}_{k=1}^K$ within $y_c$, where $K$ denotes the total number of objects. For each object, the corresponding prompt $y_k$ is recognized, and its dimensions are estimated. This includes considering the object's real-world size and its relationship with other objects to determine its relative size, facilitating the placement of all objects within the same scene.

Subsequently, LLMs sequentially position each object within the scene. In designing each object's placement, LLMs articulate the spatial relationships with relevant entities using programmable language descriptions that explicitly outline all mathematical calculations. This language is then converted into a program executed by a runtime, such as a Python interpreter, to produce the layout solution. These layouts, which include scale factors, Euler angles, and translation vectors, are employed to transform 3D Gaussians from local coordinates to global coordinates during rendering.

Furthermore, for each object $O_k$, LLMs decomposes its layout space into $n_k$ complementary regions, each with distinct attributes and different subprompts $\left\{y_{k, l}\right\}_{l=1}^{n_k}$. These complementary regions are designed to be non-overlapping and collectively encompass the entire layout space of their respective object. To generate meaningful and accurate complementary regions, LLMs employ a structured decomposition process that segments the space of object \( O_k \) into hierarchical divisions based on depth, width, and length dimensions. This process is documented using programmable language descriptions and subsequently converted into precise bounding boxes by a program. Details on the prompts used for this program-aided layout planning are provided in Appendix~\ref{supp:layout}.

\subsection{Semantic Score Distillation Sampling}
\paragraph{Prompt-Guided Semantic 3D Gaussian Representation}
To generate 3D scenes involving multiple objects with diverse attributes and to precisely control the attributes of distinct spatial regions within each object, it is essential to utilize features that represent the fine-grained semantics of 3D Gaussians. We design new prompt-guided semantic 3D Gaussian representations. During initialization, the subprompt $y_{k,l}$ corresponding to the $i$-th Gaussian is encoded via the CLIP text encoder $\Phi$~\citep{radford2021learning} to obtain the high-dimensional semantic embedding, $\mathbf{h}_i = \Phi(y_{k,l}) \in \mathbb{R}^{d_{\mathbf{h}}}$. Given the significant memory demands imposed by the large dimensions of $d_{\mathbf{h}}$, a lightweight autoencoder is employed. This autoencoder effectively compresses the scene's high-dimensional semantic embeddings into more manageable, low-dimensional representations, represented as $\mathbf{f}_i = E(\mathbf{h}_i) \in \mathbb{R}^{d_{\mathbf{f}}}$.
The loss function for the autoencoder is defined as:
\begin{equation}
     \mathcal{L}_{ae}= \sum_{i \in \mathcal{N}} d_{ae}(D(E(\mathbf{h}_i)), \mathbf{h}_i)
\end{equation}
where \( d_{ae} \) denotes the metric combining the \( \mathcal{L}_1 \) loss and the symmetric cross entropy loss from CLIP 
~\citep{radford2021learning}.

The $i$-th Gaussian is then augmented with a semantic embedding $\mathbf{f}_i \in \mathbb{R}^d$. And semantic information is integrated into the rendered 2D image by rendering the semantic embedding at pixel $v$ using the formula:
\begin{equation}
    \mathbf{F}(v) = \sum_{i \in \mathcal{N}} \mathbf{f}_i \alpha_i \prod_{j=1}^{i-1} (1 - \alpha_j)
    \label{eq:semantic_pixel}
\end{equation}
The rendered semantic embedding \(\mathbf{F}(v)\), derived from \eqref{eq:semantic_pixel}, is fed into the decoder \(D\) to reconstruct $\mathbf{S}(v) = D(\mathbf{F}(v)) \in \mathbb{R}^{d_{\mathbf{h}}}$ and then generates a semantic map \(\mathbf{S} \in \mathbb{R}^{H \times W \times d_{\mathbf{h}}}\) indicating the rendered image's semantic attributes.

\paragraph{Semantic Score Distillation Sampling}
To enable fine-grained controllable generation, the generated semantic map is integrated into the spatial composition of scores for distillation sampling. The subprompt $y_{k, l}$ is processed through the CLIP text encoder $\Phi$ to produce the subprompt embedding $\mathbf{q}_{k, l} = \Phi\left(y_{k, l}\right) \in \mathbb{R}^{d_{\mathbf{h}}}$. The probability that pixel $v$ corresponds to subprompt $y_{k, l}$ is computed as:
\begin{equation}
p(k, l \mid v)=\frac{\exp \left(\cos \left(\mathbf{q}_{k, l}, \mathbf{S}(v)\right) / \tau\right)}{\sum_{k^{\prime}=1}^K \sum_{l^{\prime}=1}^{n_{k^{\prime}}} \exp \left(\cos \left(\mathbf{q}_{k, l}, \mathbf{S}(v)\right) / \tau\right)}
\end{equation}
where $\tau$ is a temperature parameter learned by CLIP and $\cos (\cdot, \cdot)$ denotes cosine similarity. This facilitates the derivation of the mask $\mathbf{M}_{k,l}(v)$, which indicates whether the semantic properties of pixel $v$ align with subprompt $y_{k,l}$.
\begin{equation}
\mathbf{M}_{k, l}(v)= \begin{cases}1 & \text { if }(k, l)=\arg \max _{k^{\prime}, l^{\prime}} p\left(k^{\prime}, l^{\prime} \mid v\right) \\ 0 & \text { otherwise }\end{cases}
\end{equation}
The semantic mask $\mathbf{M}_{k,l} \in \{0, 1\}^{H \times W}$ is subsequently utilized to guide the score distillation sampling. To ensure that the Gaussians near the edges of objects are not overlooked, the mask \(\mathbf{M}_{k,l}\) is subjected to a max pooling operation with a \(5 \times 5\) kernel, resulting in \(\mathbf{\hat{M}}_{k,l}\). Although diffusion models generally lack an inherent distinction at the object and part levels in their latent spaces or attention maps for fine-grained control~\citep{lian2024llmgrounded}, recent advancements in compositional 2D image generation have implemented spatially-conditioned generation~\citep{chen2024training, yang2024mastering, xie2023boxdiff}. This is achieved through regional denoising or attention manipulation, allowing for fine-grained control over the semantics of the generated images. Specifically, the overall denoising score is calculated as the aggregate of the individually masked denoising scores for each visible subprompt $y_{k,l}$:
\begin{equation}
\hat{\epsilon}_\phi\left(x_t ; \mathbf{y}, t\right) = \mathbb{E}_{k,l} \left[\epsilon_\phi\left(x_t ; y_{k,l}, t\right) \odot \mathbf{\hat{M}}_{k,l}\right]
\end{equation}
where $\odot$ denotes element-wise multiplication. Instead of conditioning the diffusion models on a single text prompt, our semantic score distillation sampling employs the compositional denoising score as follows:
\begin{equation}
\nabla_\theta \mathcal{L}_{\mathrm{SemanticSDS}}=\mathbb{E}_{\epsilon, t}\left[w(t)\left(\hat{\epsilon}_\phi\left(x_t ; \mathbf{y}, t\right)-\epsilon\right) \frac{\partial \mathbf{x}}{\partial \theta}\right]
\label{eq:semanticSDSLoss}
\end{equation}
This methodology effectively leverages the expressive compositional generation capabilities of pretrained 2D diffusion models for text-to-3D generation. Further details on \method are provided in Appendix~\ref{supp:semanticSDS}.

\begin{figure}[ht]
  \centering
  \includegraphics[width=0.98\linewidth]{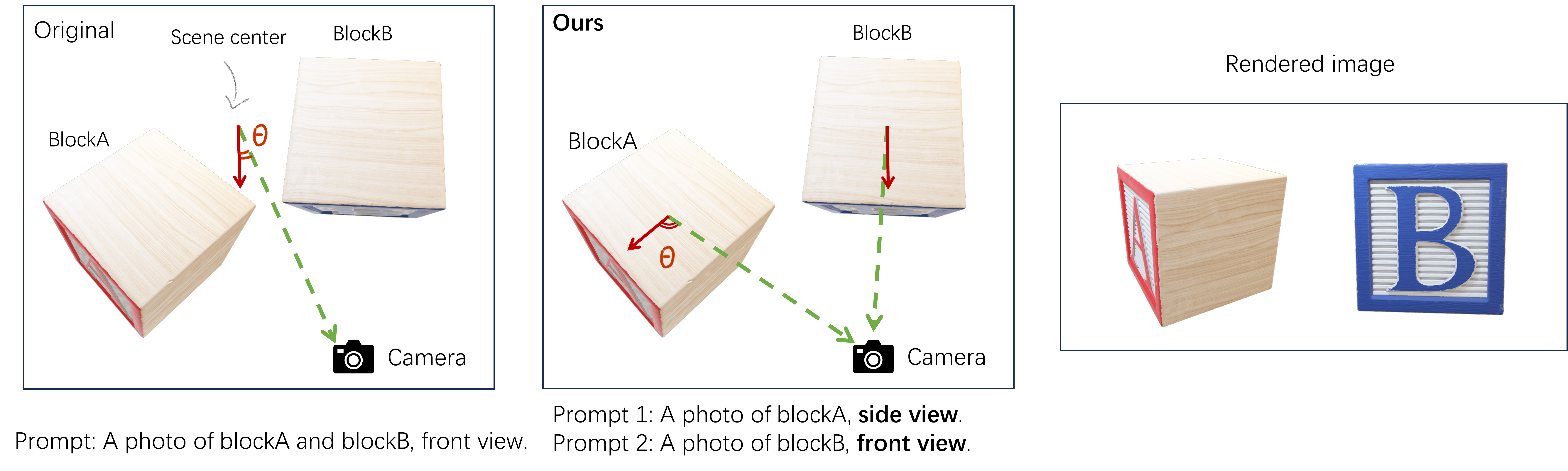}
  \caption{Illustration of our proposed object-specific view descriptor for global scene optimization.}
  \label{fig:view-descriptor}
\end{figure}

\paragraph{Object-Specific View Descriptor for Global Scene Optimization}
Unlike object-centric optimization, scenes do not exhibit distinct perspectives as individual objects do. 
Effective scene generation necessitates precise, part-level control over the optimization of distinct object views. 
Terms such as "side view" or "back view" are rarely applicable to multi-object scenes, and pretrained diffusion models often struggle to generate images accurately from such prompts~\citep{li2023gligen}. Moreover, within a single rendered image, different objects may be visible from varying perspectives. Using a unified view descriptor for an entire scene with multiple objects exacerbates the Janus Problem~\citep{poole2023dreamfusion}. Although the compositional optimization scheme alternates between local object optimizations and global scene optimizations~\citep{zhou2024gala3d}, allowing for the correct optimization of different views of objects in local coordinates, it is confounded by optimizations under global coordinates. This limits the frequency of global scene optimizations and results in a lack of scene coherence, harmony, and lighting consistency.

To address this issue, in our \method, we append an object-specific view descriptor \(y^{\text{view}}_{k}\) to the corresponding subprompts \(\{y_{k,l}\}_{l=1}^{n_K}\) to optimize individual objects within the rendered image (in Figure~\ref{fig:view-descriptor}). The same view descriptor \(y^{\text{view}}_{k}\) is consistently applied across different parts of each multi-attribute object. Specifically, we determine the camera's elevation and azimuth angles relative to each object by computing the angle between the vector \(\hat{n}\), which extends from the object to the camera, and specific reference axis vectors, such as the positive z-axis. This calculation facilitates the selection of the most appropriate object-specific view descriptor. For instance, if the angle between \(\hat{n}\) and the positive \(z\)-axis remains below a predefined threshold, indicative of a high azimuth angle, the descriptor \(y^{\text{view}}_{k}\) is assigned as an overhead view descriptor for that object.

\section{Experiments}
\paragraph{Implementation Details.}
The guidance model is implemented using the publicly accessible diffusion model, StableDiffusion~\citep{rombach2022high}, specifically utilizing the checkpoint \textit{runwayml/stable-diffusion-v1-5}. Positions of the Gaussians are initialized using Shap-E~\citep{jun2023shap}, with each object initially comprising 12288 Gaussians. For densification, Gaussians are cloned or split based on the view-space position gradient using a threshold \( T_{\text{pos}} = 2 \), with semantic embeddings copied. Compactness-based densification is also applied every 2000 iterations, involving each Gaussian and one of its nearest neighbors, as described in GSGEN~\citep{chen2024text}. Pruning involves removing Gaussians with opacity lower than \( \alpha_{\min} = 0.3 \), as well as those with excessively large radii in either world-space or view-space, every 200 iterations.

Training alternates between local and global optimization. During global optimization, the rendered objects vary by switching between the entire scene and pairs of objects. Camera sampling maintains the same focal length, elevation, and azimuth range as specified in \citep{chen2024text}. The threshold for selecting object-specific view descriptors includes: an overhead view descriptor for elevation angles exceeding 60°, a front view descriptor for azimuth angles within ±45° of the positive x-axis, and a back view descriptor for ±45° angles on the negative x-axis.

\begin{table}[htbp]
    \centering
    \caption{Quantitative Comparison}
    \resizebox{\textwidth}{!}{%
    \begin{tabular}{@{}lccccc@{}}
    \toprule
    Metrics & GraphDreamer & GSGEN & LucidDreamer & GALA3D& SemanticSDS (Ours) \\
    \midrule
    CLIP Score $\uparrow$ & 0.289 & 0.314 & 0.311 & 0.305 & \textbf{0.321} \\
    Prompt Alignment $\uparrow$ & 56.9 & 63.3 & 64.4 & 85.0 & \textbf{91.1} \\
    Spatial Arrangement $\uparrow$ & 53.8 & 62.8 & 65 & 80.0 & \textbf{85.7} \\
    Geometric Fidelity $\uparrow$ & 53.8 & 71.1 & 71.8 & 80.3 & \textbf{83.0} \\
    Scene Quality $\uparrow$ & 54.9 & 71.2 & 65.9 & 82.3 & \textbf{86.9} \\
    \bottomrule
    \end{tabular}
    }
    \label{tab:quant}
\end{table}

\paragraph{Baseline methods.}
To evaluate the performance of \method on the complex Text-to-3D task involving multiple objects with varied attributes, we compare it with state-of-the-art (SOTA) methods. These include the compositional 3D generation method GALA3D~\citep{zhou2024gala3d} and GraphDreamer~\citep{gao2024graphdreamer}, noted for their ability to generate intricate scenes with multiple objects. Additionally, we consider GSGEN~\citep{chen2024text} and LucidDreamer~\citep{liang2024luciddreamer}, both are capable of producing high-quality, complex objects with diverse attributes.

\begin{figure}[t]
  \centering
  \includegraphics[width=0.99\linewidth]{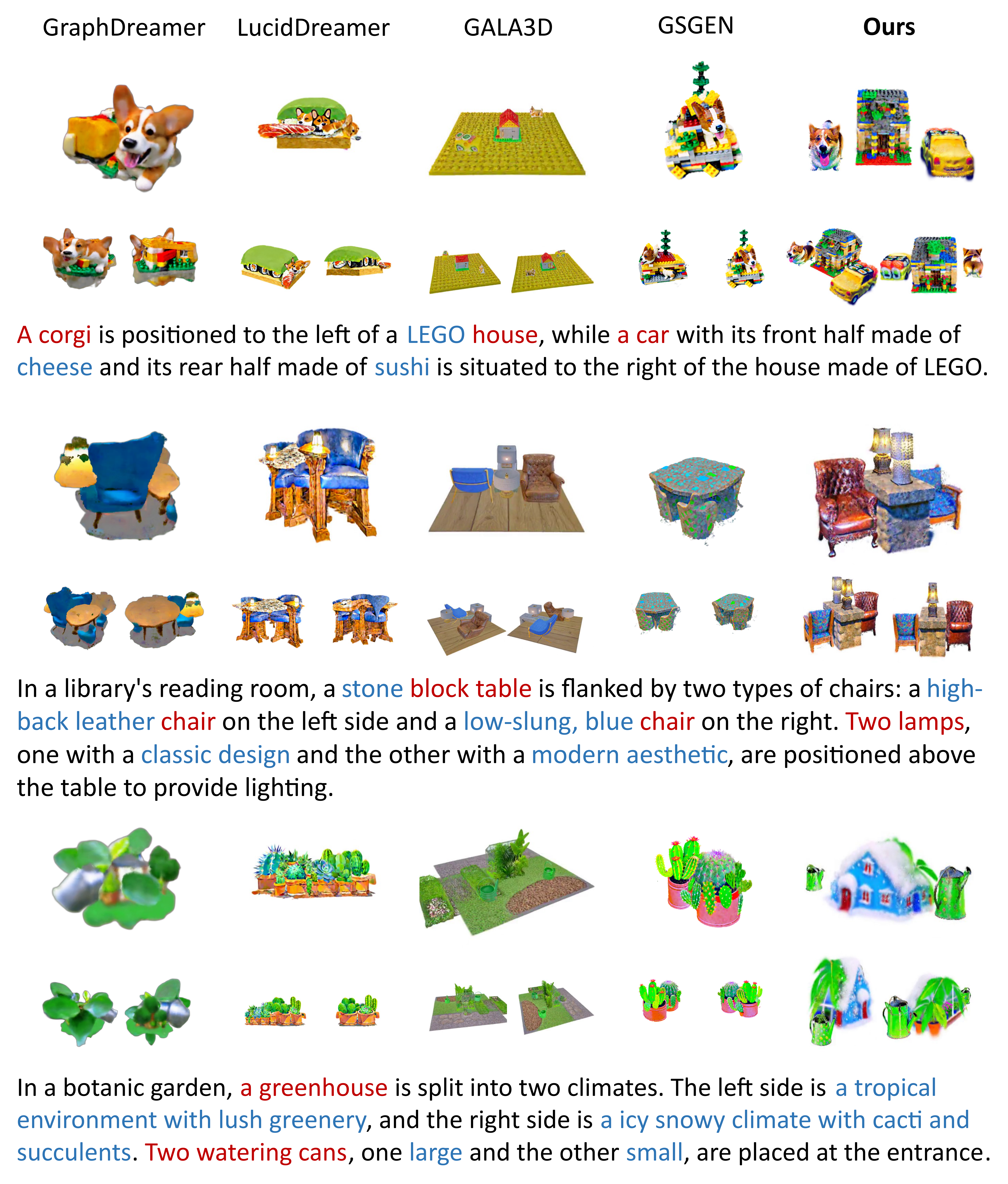}
  \caption{\textbf{Qualitative comparisons of text-to-3D generation.} Comparison results demonstrate that \method synthesizes more precise and realistic multi-object scenes with better visual details, geometric expressiveness, and semantic consistency.}
  \label{fig:main_results}
  \vspace{-4mm}
\end{figure}

\paragraph{Metrics.} 

CLIP Score~\citep{radford2021learning} is employed as the evaluation metric to assess the quality and consistency of the generated 3D scenes with textual descriptions. However, CLIP tends to focus on the primary objects within the rendered image, and when used to evaluate complex text-to-3D tasks involving multiple objects with varied attributes, it may not adequately assess the geometry of all objects or the rationality of their spatial arrangements. This limitation results in a misalignment with human judgment regarding evaluation criteria. Therefore, following~\citet{wu2024gpt}, GPT-4V is utilized as a human-aligned evaluator to compare 3D assets based on predefined criteria. These criteria include: (1) Prompt Alignment: ensuring that all objects specified in the user prompts are present and correctly quantified; (2) Spatial Arrangement: evaluating the logical and thematic spatial arrangement of objects; (3) Geometric Fidelity: assessing the geometric fidelity of each object for realistic representation; and (4) Scene Quality: determining the overall scene quality in terms of coherence and visual harmony. More details on metrics are provided in the Appendix~\ref{supp:metrics}.

\subsection{Main Results}
\paragraph{Quantitative Analysis}
To evaluate the performance of \method in Text-to-3D tasks involving multiple objects with varied attributes, quantitative metrics were employed. As shown in Table~\ref{tab:quant}, the CLIP Score indicates that \method exhibits strong alignment with the primary semantics of user prompts. Specifically, \method excels in Prompt Alignment, ensuring that all objects specified in user prompts are present and correctly quantified. Additionally, it demonstrates superior performance in Spatial Arrangement, effectively designing the layout of interactive objects to support the scene's intended theme. Furthermore, by explicitly guiding SDS with rendered semantic maps, \method achieves outstanding generation of individual objects with diverse attributes across different spatial components, resulting in high scores in object-level Geometric Fidelity. Additionally, the use of compositional 3D Gaussian Splatting for scene representation helps \method to effectively disentangle objects within the scene. This, combined with explicit semantic guidance to the SDS, contributes to achieving the highest score in Scene Quality.

\paragraph{Qualitative Analysis}

To intuitively demonstrate the superiority of the proposed method in generating complex 3D scenes with multiple objects possessing diverse attributes, a qualitative comparison with baseline models is conducted. As illustrated in Figure~\ref{fig:main_results}, GALA3D, with a compositional optimization scheme, successfully generates individual objects that align with user prompts. However, it fails to produce plausible results when objects have multiple attributes. Although GSGEN and LucidDreamer generate high-quality individual objects, the presence of multiple objects often leads to entanglement, compromising consistency with user prompts. Additionally, these models are unable to generate reasonable objects when individual objects possess numerous attributes. In contrast, \method employs guided diffusion models with explicit semantics, effectively generating scenes that include multiple objects with diverse attributes. Moreover, by utilizing program-aided layout planning, \method produces more coherent layouts than GALA3D in scenarios involving complex spatial relationships among multiple objects. For example, in Figure~\ref{fig:teaser}, both table lamps are correctly placed on the table without appearing to float when using \method.

\begin{wrapfigure}{r}{0.40\textwidth}
\vspace{-10mm}
  \begin{center}
    \includegraphics[width=0.35\textwidth]{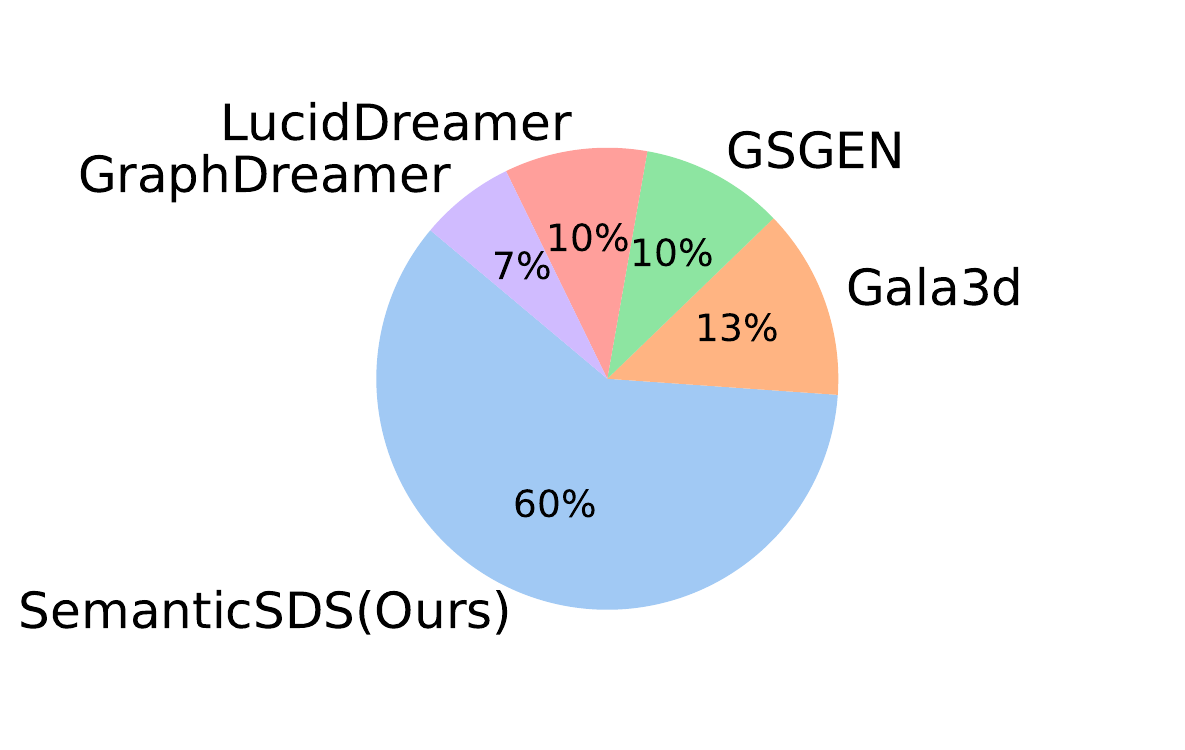}
  \end{center}
  \vspace{-3.5mm}
   \caption{\textbf{User study results.} \method~is  preferred 60\% of the time by users than baseline methods.}
  \label{fig:user_pie}
\vspace{-12mm}
\end{wrapfigure}
\paragraph{User Study} We conducted a user study to compare our method with baseline methods across 30 scenes involving more than 100 objects. Each participant was shown a user prompt alongside 3D scenes generated by all methods simultaneously and asked to select the most realistic assets based on geometry, prompt alignment, and accurate placement. Figure~\ref{fig:user_pie} illustrates that \method significantly outperformed previous methods in terms of human preference.

\begin{figure}[ht]
  \centering
  \vspace{-0.2in}
  \includegraphics[width=0.9\linewidth]{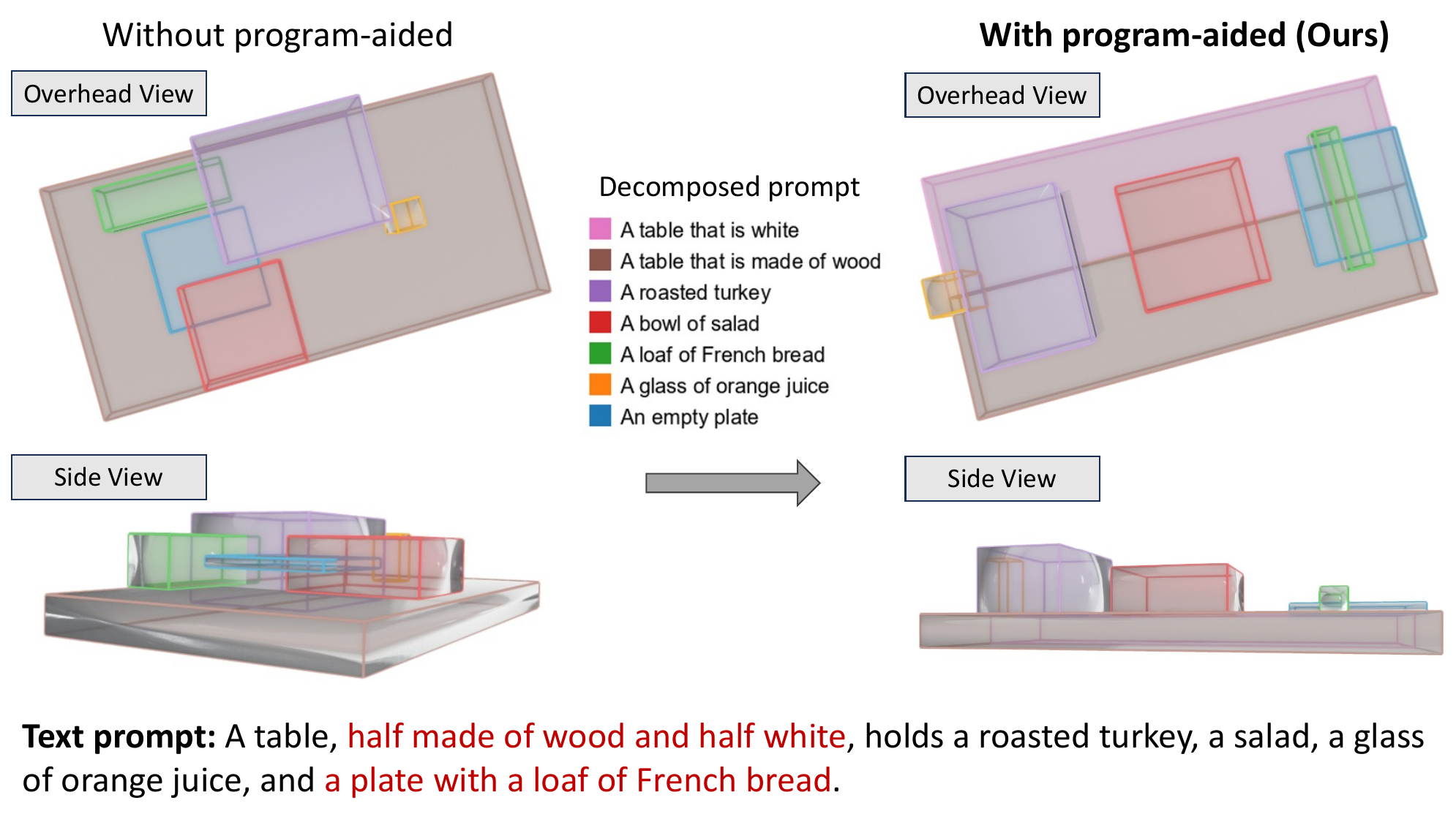}
  \caption{Qualitative comparisons between without and with our program-aided layout planning.}
  \label{fig:layout}
\end{figure}
\subsection{Model Analysis}


\paragraph{Effectiveness of Program-aided Layout Planning}

We assess the necessity of program-aided layout planning through an ablation study. The qualitative comparison of generated layouts is illustrated in Figure~\ref{fig:layout}. Without program-aided planning, layout placement often lacks rationale and results in poor spatial arrangements. In contrast, the program-aided strategy positions the layouts logically and divides the layout into meaningful and precise complementary regions for objects with multiple attributes, resulting in an effective spatial arrangement.

\begin{figure}[h]
  \centering
  \includegraphics[width=0.98\linewidth]{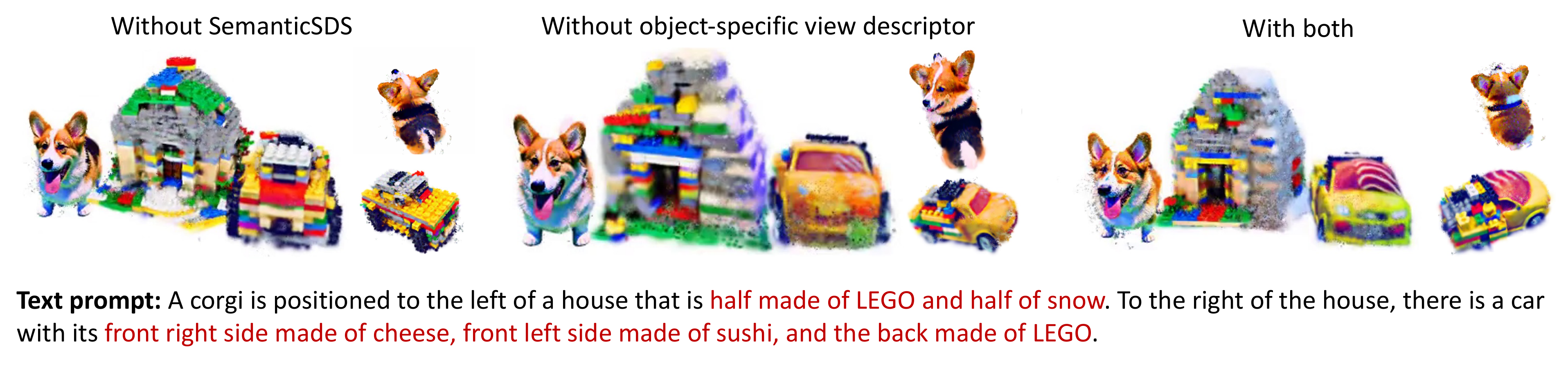}
  \caption{Qualitative analysis. Our \method provides more precise and fine-grained control and our proposed object-specific view descriptor helps with better multi-view understanding.}
  \label{fig:model_analysis}
  \vspace{-4mm}
\end{figure}

\paragraph{Impact of Semantic Score Distillation Sampling}
Ablation experiments are performed on Semantic Score Distillation Sampling to evaluate the effects of explicitly guiding SDS with rendered semantic maps. In Figure~\ref{fig:model_analysis}, without \method, while objects with single attributes are generated effectively, those with varied attributes often experience blending issues. For instance, the "house" shows snow bricks mixed with LEGO bricks, failing to meet the user prompt's spatial requirements. The snow bricks are inaccurately represented as white LEGO bricks, which do not align with the intended attributes. Additionally, one attribute may dominate, causing others to disappear, such as in the "car" with three attributes in Figure~\ref{fig:model_analysis}. Conversely, SemanticSDS enables precise control over the attributes in distinct spatial regions of each object, producing objects with diverse attributes and smooth transitions between regions with different attributes.

\paragraph{Object-Specific View Descriptor}
To assess the effectiveness of the object-specific view descriptor, we replace it with the scene-centric view descriptor utilized by GSGEN during global optimization. This change increases the occurrence of the Janus Problem, as illustrated by the overhead view of the corgi in the middle of Figure~\ref{fig:model_analysis}. These findings highlight the crucial role of selecting an appropriate view descriptor to enhance the plausibility of generated 3D scenes.

\section{Conclusion}
In this paper, we introduce \method, a novel SDS method that significantly enhances the expressiveness and precision of compositional text-to-3D generation. By leveraging program-aided layout planning, semantic embeddings, and explicit semantic guidance, we unlock the compositional priors of pre-trained diffusion models and achieve realistic high-quality generation in complex scenarios.
Our extensive experiments demonstrate that \method achieves state-of-the-art results for generating complex 3D content. As we look to the future, we envision \method as a foundation for even more applications, such as automatic editing and closed-loop refinement, paving the way for unprecedented levels of creativity and innovation in 3D content generation.


\bibliography{iclr2024_conference}
\bibliographystyle{iclr2024_conference}
\clearpage
\appendix
\section{More Implementation Details}
\subsection{Prompts for Program-aided Layout Planning}
\label{supp:layout}

\begin{figure}[ht]
  \centering
  \includegraphics[width=0.99\linewidth]{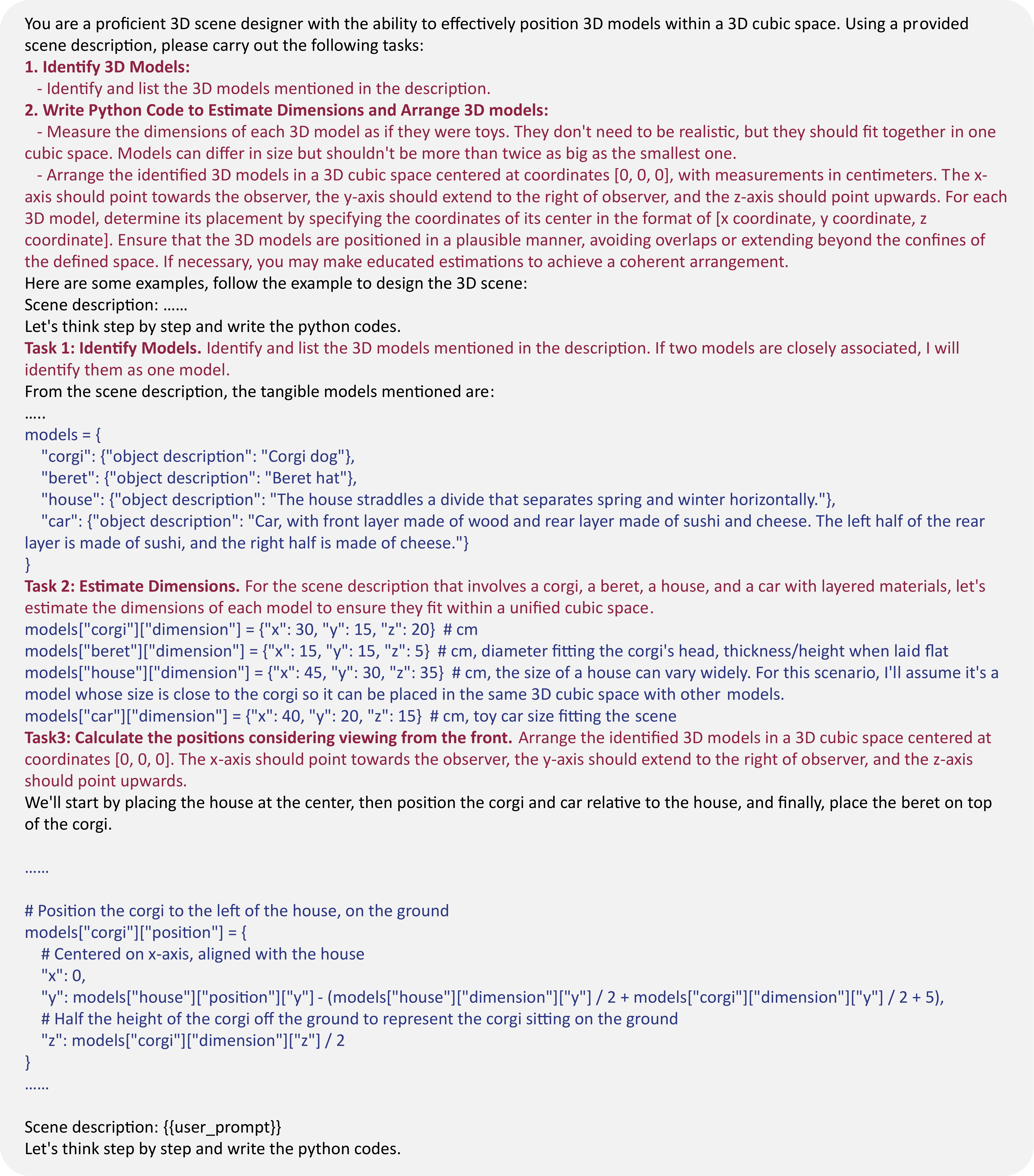}
  \caption{The prompt for scene-level decomposition in program-aided layout planning.}
  \label{fig:scene_decompose}
  \vspace{-4mm}
\end{figure}

\begin{figure}[ht]
  \centering
  \includegraphics[width=0.99\linewidth]{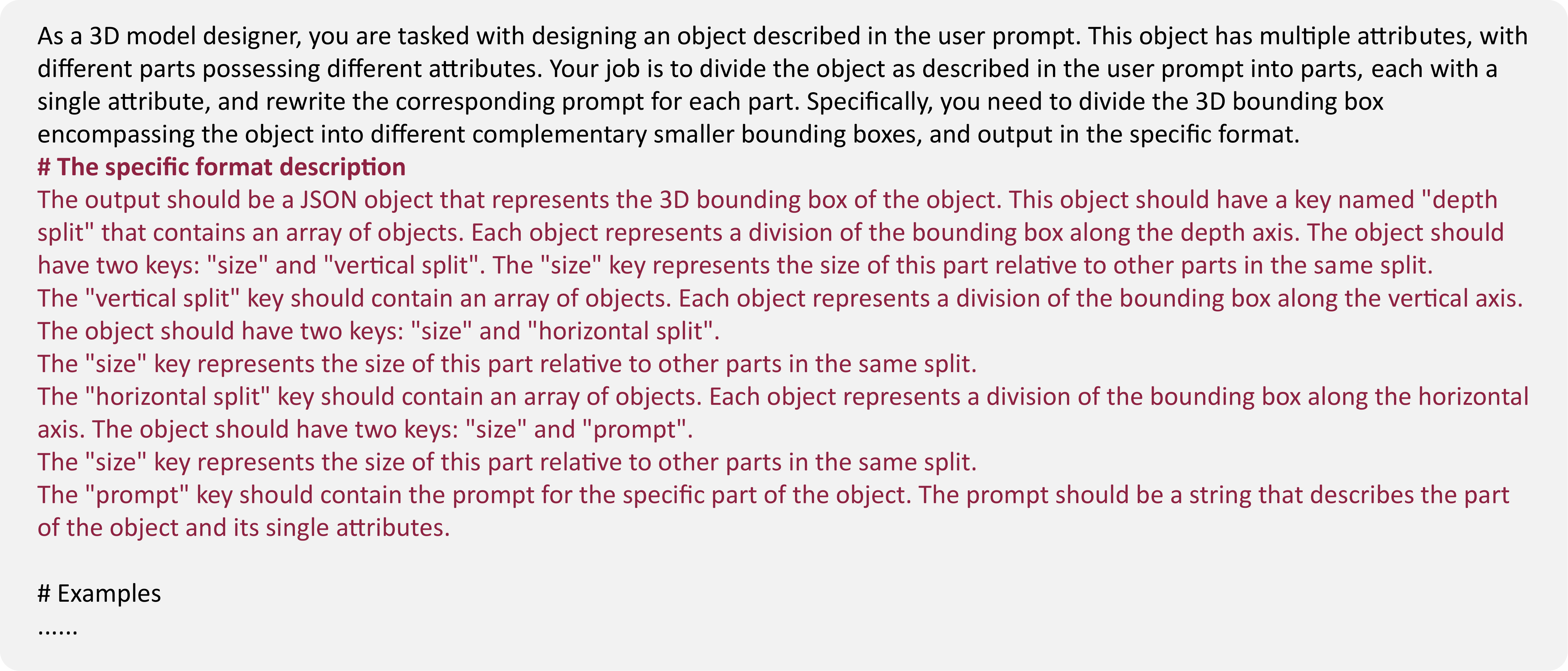}
  \caption{The prompt for decomposing each object into complementary regions.}
  \label{fig:object_decompose}
  \vspace{-4mm}
\end{figure}

Large Language Models (LLMs) have the potential for spatial awareness; however, precise 3D layout generation from vague language descriptions is challenging. This difficulty arises because 3D digital data and corresponding natural language descriptions often do not appear simultaneously~\citep{hong20233d, xu2023pointllm}. Moreover, minor numerical changes, which might not be reflected in imprecise language, can lead to unrealistic spatial arrangements of 3D scenes. Additionally, the spatial arrangement of multi-object scenes requires numerous parameters, making a program-aided approach necessary to bridge the gap between natural language descriptions and 3D digital data.

Specifically, we decompose the process of generating multiple objects with diverse attributes into two steps: scene-level decomposition and object-level decomposition. In scene decomposition, we guide LLMs to translate user prompts into Python programs, using explicit mathematical operations to represent relationships between objects. For object decomposition, since complementary regions are designed to be non-overlapping and collectively encompass the entire layout space of their respective objects, we devised a scheme employing structured JavaScript Object Notation~(JSON) to represent hierarchical divisions based on depth, width, and length dimensions. Figures~\ref{fig:scene_decompose} and~\ref{fig:object_decompose} illustrate the detailed prompts for scene and object decomposition, respectively.

\subsection{SemanticSDS}
\label{supp:semanticSDS}

\paragraph{Camera Sampling} Training alternates between local and global optimization. During local optimization, objects are not transformed into global coordinates. In global optimization, the rendering of objects varies by switching between the entire scene and pairs of objects to better optimize those that interact or occlude each other. When rendering only a pair of objects, the camera's look-at point is sampled at the midpoint between the two objects rather than the center of the entire scene. Additionally, we apply a dynamic camera distance from the object pair to ensure the objects are appropriately sized in the rendered images. Specifically, the camera distance is determined by the scale of the objects and the distance between their centers.

\paragraph{Pooling of Semantic Masks} Given that the rendered RGB images and the semantic map have sizes of $512 \times 512$, whereas the latents for denoising are of size $64 \times 64$, we convert the semantic map $\mathbf{S}$ into masks to compose the denoising scores predicted by diffusion models. Subsequently, for each mask $\mathbf{M}_{k,l} \in \{0, 1\}^{512 \times 512}$, we apply average pooling with a stride of 8 using an $8 \times 8$ kernel to downsample the data. To ensure that Gaussians near the edges of objects and isolated Gaussians are not overlooked, the mask $\mathbf{M}_{k,l}$ undergoes a max pooling operation with a $5 \times 5$ kernel, resulting in $\mathbf{\hat{M}}_{k,l}$.

\paragraph{Compositional Optimization Scheme} The compositional optimization scheme encompasses both global scene and local object optimizations. Only global scene optimizations apply affine transformations to convert objects from local to global coordinates. During local optimization, \(\theta\) in \eqref{eq:semanticSDSLoss} includes the mean, covariance, and color of individual Gaussians. In global scene optimization, \(\theta\) additionally includes the parameters of affine transformations—translation, scale, and rotation—that convert local to global coordinates. 

\subsection{Details of Metrics}
\label{supp:metrics}

\begin{figure}[ht]
  \centering
  \includegraphics[width=0.99\linewidth]{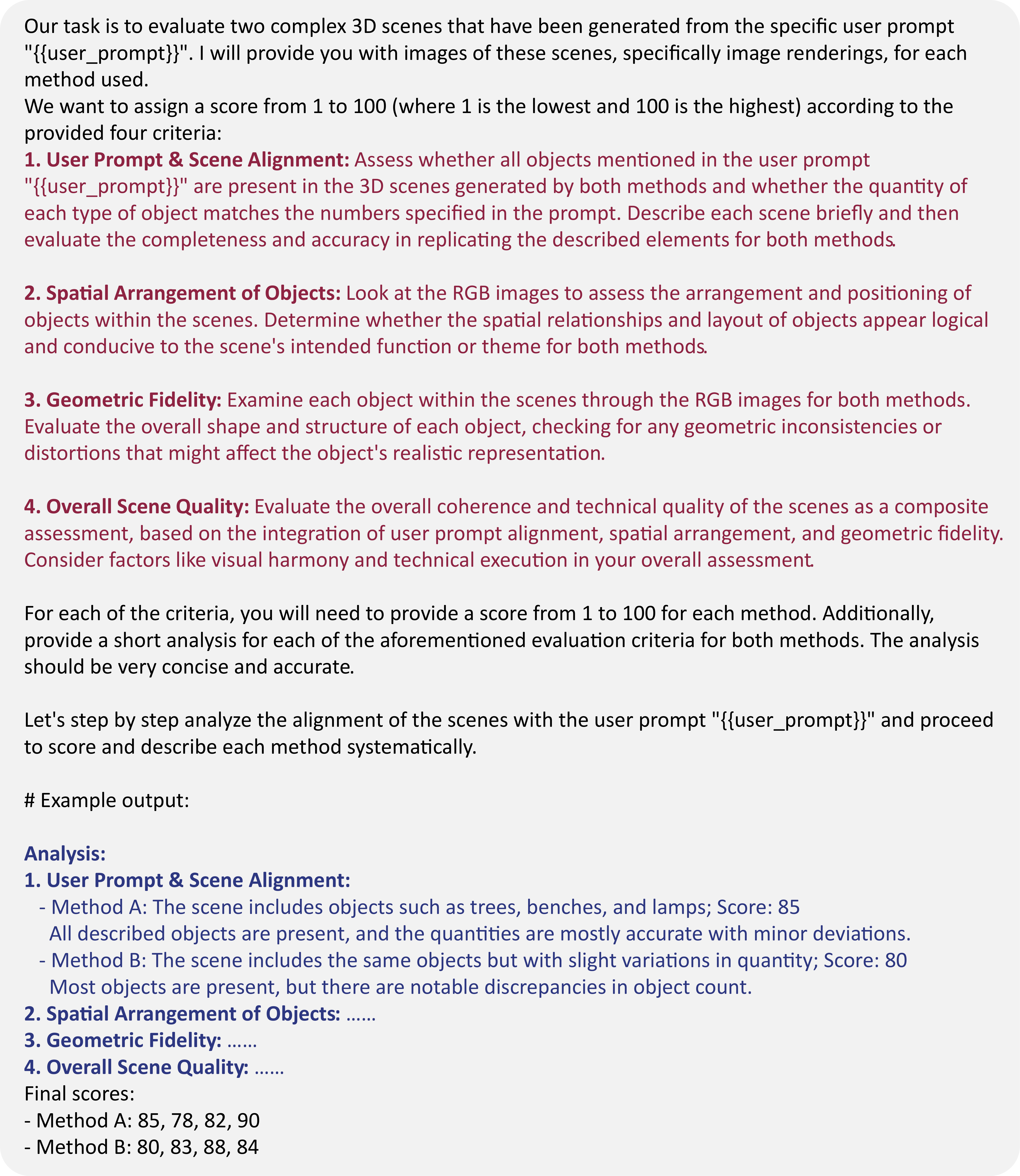}
  \caption{The prompt for guiding GPT-4 as a human-aligned evaluator}
  \label{fig:gpt4eval}
  \vspace{-4mm}
\end{figure}

\paragraph{CLIP Score} The CLIP score utilizes CLIP embeddings \citep{radford2021learning} to evaluate text-to-3D alignment. Following previous methods \citep{zhou2024gala3d, gao2024graphdreamer}, we calculate the cosine similarity between the user prompt and scene images rendered from different perspectives. For each scene, we take the maximum CLIP score from all rendered images as the representative score. We then compare the average of these maximum scores across different scenes for each method. 

\paragraph{GPT-4V as A Human-Aligned Evaluator} Due to the limitations of the CLIP score in capturing spatial arrangement and geometric fidelity, we follow \citet{wu2024gpt} and employ GPT-4V to evaluate complex 3D scenes involving multiple objects with varied attributes. Specifically, we provide GPT-4V with rendered images of the same 3D scene generated by different methods and require it to score each scene on four aspects: Prompt Alignment, Spatial Arrangement, Geometric Fidelity, and Scene Quality, each on a scale from 1 to 100. For each scene and method pair, we perform three independent evaluations. The final score for each method is obtained by averaging the scores across different scenes and comparisons with other methods. Figure~\ref{fig:gpt4eval} presents the prompt used to guide the GPT-4V evaluator. In the prompt, "method A" and "method B" are used to anonymize the methods, preventing name bias in GPT-4V's judgment.

\section{More Synthesis Results}

\begin{figure}[ht]
  \centering
  \vspace{-0.3in}
  \includegraphics[width=0.9\linewidth]{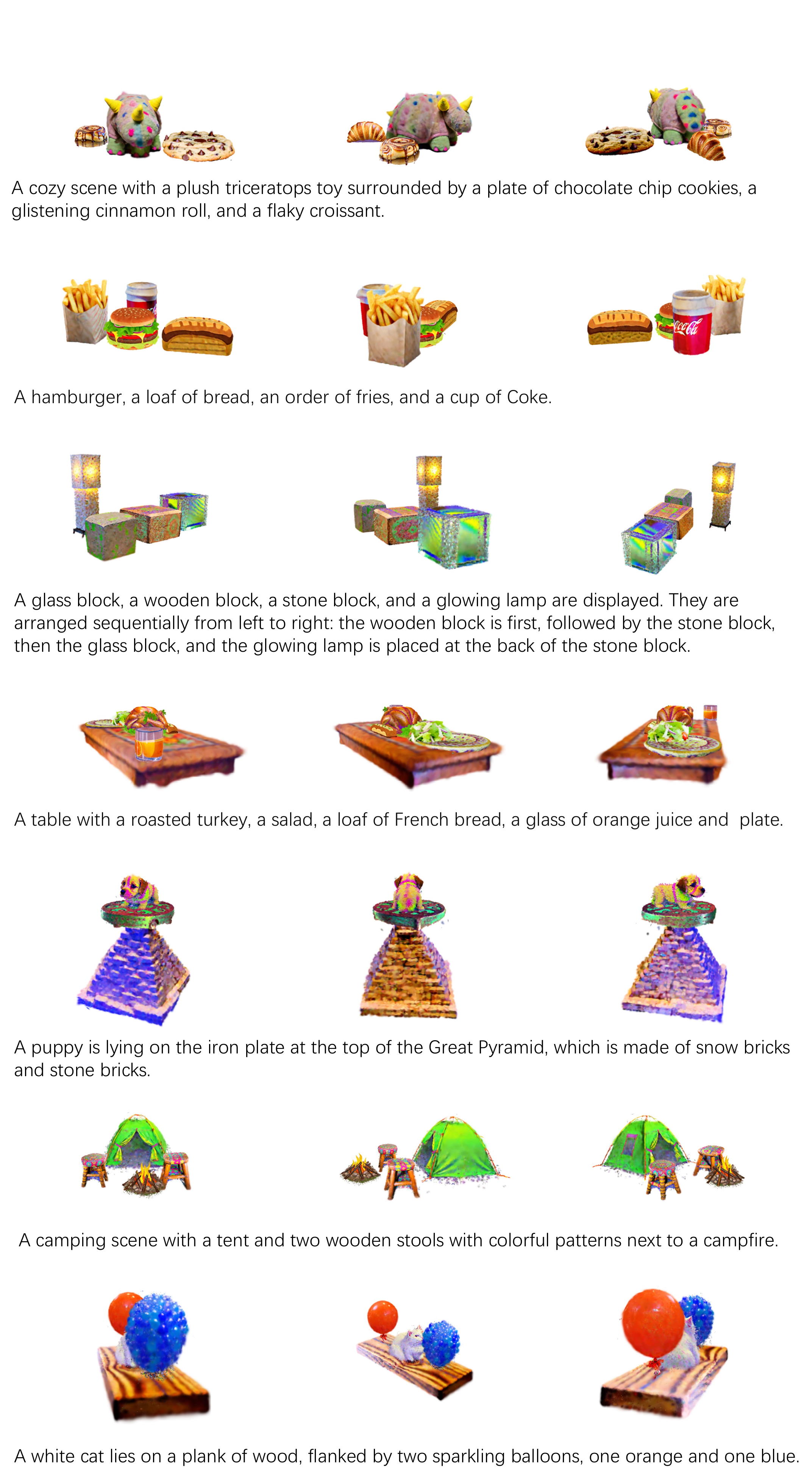}
  \caption{More synthesis results of multiple objects with our \method.}
  \label{fig:app-multi}
  \vspace{-0.2in}
\end{figure}

\begin{figure}[ht]
  \centering
  \vspace{-0.4in}
  \includegraphics[width=0.9\linewidth]{figs/app-single.pdf}
  \caption{More synthesis results of single object with diverse attributes with our \method.}
  \label{fig:app-single}
  \vspace{-0.2in}
\end{figure}

\end{document}

%% file: math_commands.tex
\usepackage{amsmath,amsfonts,bm}

\def\eqref#1{equation~\ref{#1}}

\def\1{\bm{1}}

\DeclareMathAlphabet{\mathsfit}{\encodingdefault}{\sfdefault}{m}{sl}
\SetMathAlphabet{\mathsfit}{bold}{\encodingdefault}{\sfdefault}{bx}{n}

